\documentclass{tlpfinal}

\usepackage{general}
\usepackage{graphicx}

\newcommand{\spec}{\emph{spec}}
\newcommand{\etime}{\mbox{ETIME}}
\newcommand{\netime}{\mbox{NETIME}}
\newcommand{\nnetime}{\mbox{(N)ETIME}}
\newcommand{\KB}{\emph{KB}}

\newcommand{\wnKB}{W_{\KB,n}}

\newcommand{\PnKB}{P_{\KB,n}}
\newcommand{\PnKBpr}{P_{\KB',n}}
\newcommand{\PnM}{P_{M,n}}
\newcommand{\PnMpr}{P_{M',n}}

\newcommand{\PInf}{\emph{PI}}
\newcommand{\phirskol}{\phi^{\emph{R-Skol}}}
\newcommand{\wformeq}[2]{#1\hspace{2.5mm}:\hspace{2.5mm}#2}
\newcommand{\wformtxt}[2]{#1\hspace{1mm}:\hspace{1mm}#2}

\begin{document}
\bibliographystyle{acmtrans}

\long\def\comment#1{}

\title{Lower complexity bounds for lifted inference}

\author[M. Jaeger]
{Manfred Jaeger \\
Aalborg University\\
E-mail: jaeger@cs.aau.dk
}


\maketitle

\label{firstpage}

\begin{abstract}

One of the big challenges in the development of  probabilistic relational (or probabilistic
logical) modeling and learning frameworks is the design of inference techniques 
that operate on the  level of the abstract model representation language, rather 
than on the level of ground, propositional instances of the model.     
Numerous approaches for such ``lifted inference'' techniques have been proposed.
While it has been demonstrated that these techniques will lead to  significantly
more efficient inference on some specific models, there are only very recent and
still quite restricted results that show the feasibility of lifted inference on certain
syntactically defined classes of models. Lower complexity bounds that imply some
limitations for the feasibility of lifted inference on more expressive model classes 
were established earlier in (Jaeger 2000). However, it is not immediate that these results
also  apply  
to the type of modeling languages that currently receive the most attention, i.e.,  weighted, 
quantifier-free formulas. In this paper we extend these earlier results, 
and show that under the assumption that NETIME$\neq$ETIME, there is no polynomial lifted 
inference algorithm for knowledge bases of weighted, quantifier- and function-free 
formulas. Further strengthening earlier results, this is also shown to hold for 
approximate inference, and for knowledge bases not containing the equality predicate. 
\end{abstract}


\section{Introduction}

Probabilistic logic models (a.k.a. probabilistic or statistic relational models) provide
high-level representation languages for probabilistic models of structured 
data \cite{Breese92,Poole93,Sato95,NgoHadHel95,Jaeger97UAI,FriGetKolPfe99,KerDeR01,MilMarRusSonOngKol05,VenDenBru06,TasAbbKol02,RicDom06}.
While supporting model specifications at an abstract, first-order logic level, inference is 
typically performed at the level of concrete ground instances of the models, i.e., at the 
propositional level. This mismatch between model specification and inference methods has been 
noted early on~\cite{Jaeger97UAI}, and has given rise to numerous proposals for  inference techniques that 
operate at the high level of the underlying model specifications~\cite{Poole03,BraAmiRot05,MilZetKerHaiKae08,KisPoo09,JhaGogMelSuc10,GogDom11,BroTagMeeDavDeR11,Broeck11,FieBroThoGutDeR11}. Inference methods of this 
nature have collectively become known as ``lifted'' inference techniques.

The concept of lifted inference is mostly introduced on an informal level: 
``\emph{...lifted, that is,
deals with groups of random variables at a first-order level}''~\cite{BraAmiRot05}; 
``\emph{The act of exploiting the high level structure in
relational models is called lifted inference}''~\cite{ApsBra11}; 
``\emph{The idea behind lifted inference is to carry out
as much inference as possible without propositionalizing}~\cite{KisPoo09};
``\emph{lifted inference, which deals with groups of indistinguishable
variables, rather than individual ground atoms}~\cite{SinNatDom10}. While, thus, the 
term lifted 
inference emerges as a quite coherent algorithmic metaphor, it is not immediately obvious
what its exact technical meaning should be. Since quite a variety of different algorithmic 
approaches are collected under the label ``lifted'', and since most of them can degenerate 
for certain models to ground, or propositional, inference, it is difficult to precisely 
define the class of lifted inference techniques in terms of specific algorithmic 
techniques employed.

\begin{figure}
  \centering
  \includegraphics[scale=0.3]{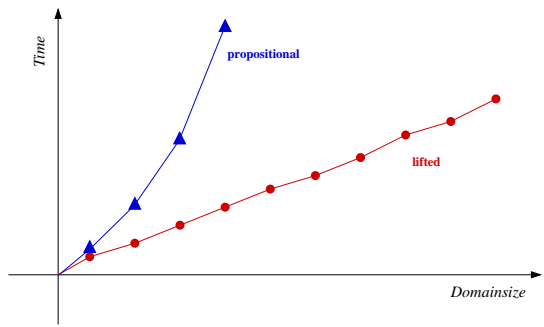}
  \caption{A typical performance evaluation}
  \label{fig:plot}
\end{figure}

A more fruitful approach is to make more precise the concept of lifted inference in terms of 
its objectives. Here one observes that
lifted inference techniques very consistently are evaluated on, and compared against each
other, by how well inference complexity scales as a function of the domain (or population)
for which the general model is instantiated. Thus, empirical evaluations of lifted 
inference techniques are usually presented in the form of domainsize vs. inference time
plots as shown in Figure~\ref{fig:plot}. 
 
Van den Broeck~\citeyear{Broeck11}, therefore, has proposed a formal definition of 
\emph{domain lifted inference} in terms of polynomial time complexity 
in the domainsize parameter. Experimental and theoretical analyses of 
existing lifted inference techniques then show that they provide domain lifted
inference in some cases where basic propositional inference techniques would 
exhibit exponential complexity (as illustrated in  Figure~\ref{fig:plot}). 
However, until recently, these positive results
were mostly limited to examples of individual models, and little was known about
the feasibility of lifted inference for certain well-defined classes of models.
First results that show the feasibility of lifted inference for whole classes
of models are given by Van den Broeck~\citeyear{Broeck11}, and Domingos and Webb~\citeyear{DomWeb12}.
  
On the other hand, \cite{Jaeger00} has shown that under certain assumptions on the 
expressivity of the modeling language, probabilistic inference is not polynomial 
in the domainsize, thereby demonstrating some inherent limitations in terms of 
worst-case complexity for the goals
of lifted inference. 
However, the results of \cite{Jaeger00} are based on types of probabilistic logic 
models that are somewhat different from the models that presently receive the most
attention: first, they essentially assume a \emph{directed} modeling framework,
in which the model represents a generative stochastic process for sampling 
relational structures. The model  is defined by specifying marginal and conditional 
probability distributions for random variables corresponding to ground atoms.
Ground instances of the model, then, can be represented by directed graphical 
models, i.e., Bayesian networks. 
While the majority of existing model classes fall into the category of 
directed models~\cite{Breese92,Poole93,Sato95,NgoHadHel95,Jaeger97UAI,FriGetKolPfe99,KerDeR01,MilMarRusSonOngKol05,VenDenBru06}, there is currently a lot of interest in \emph{undirected} models 
that are given by a set of soft constraints on relational structures, specified in 
the form of potential functions, and in the ground case giving rise to undirected 
graphical models, i.e., Markov networks. 
Secondly, the results of \cite{Jaeger00} require quite strong assumptions on the 
expressivity of the probabilistic-logic modeling language, which is required to 
allow that conditional  distributions of atoms can be specified dependent  
on unrestricted first-order properties. Much current work, in contrast, is concerned 
with languages that only incorporate certain weak fragments of first-order logic.

In this paper the general 
approach of \cite{Jaeger00}
is extended to obtain lower complexity bounds for inference
in probabilistic-logic model classes that have emer\-ged as the focus 
of interest for lifted inference techniques, i.e., undirected models 
based on quantifier- and function-free fragments of first-order logic. 

In a sharp contrast with \cite{Jaeger00}, where a ``trivial'' constant-time approximate 
inference method was described, we show that our lower complexity bounds also hold 
for approximate inference. Further sharpening earlier results, we finally establish that
the lower complexity bounds also hold for models not using the equality predicate, which 
in \cite{Jaeger00} was conjectured to be the key source of inherent complexity. 

A preliminary version of this paper has been published as~\cite{arxiv12}. Its main 
results were also already included in the survey paper~\cite{JaeBro12}, which contains
a systematic overview of known results and open problems related to the complexity
of lifted inference.

In the following section we introduce a general framework in which classes of undirected
probabilistic-logic models, and classes of associated inference problems can be 
defined. Section~\ref{sec:spectra}
reviews classic  results relating first-order logic models to the complexity 
class NETIME. Section~\ref{sec:results} contains our main results, and Section~\ref{sec:discuss}
discusses some notable differences that emerge between the results for directed and
for undirected models.

\section{Weighted Feature Models}

Similarly as \cite{RicDom06},
\cite{BroTagMeeDavDeR11} and \cite{GogDom11}
we assume the following framework: a  model, or knowledge base,  is 
given by a set of weighted formulas:
\begin{equation}
  \label{eq:kb}
\KB: \hspace{5mm}
  \begin{array}{lll}
    \phi_1(\boldv_1) & : &  w_1 \\
    \phi_2(\boldv_2) & : & w_2 \\
    \ldots &   & \ldots \\
    \phi_n(\boldv_N) & : & w_N
  \end{array}
\end{equation}
where the $\phi_i$\ are formulas in first-order predicate logic,
$w_i\in\Rset$\ are non-negative \emph{weights}, and $\boldv_i=(v_{i,1},\ldots,v_{i,k_i})$\ 
are the free variables of $\phi_i$. The case $k_i=0$, i.e., $\phi_i$\ is a sentence 
without free variables, is also permitted. The $\phi_i$\ use a given 
signature  $S$\ of relation-, function-, and constant symbols. 

An interpretation (or possible world) $(D,I)$\ for $S$\ consists of a domain $D$, and an 
interpretation function $I$\ that maps the symbols in $S$\ to functions, relations and
elements on $D$.  For a tuple $\boldd\in D^{k_i}$\ then the truth value of 
$\phi_i(  \boldv_i/\boldd  )$\ is defined, and we write 
$(D,I)\models \phi_i(\boldd)$, or simpler $I\models \phi_i(\boldd)$, if
$\phi_i( /\ \boldv_i/\boldd  )$\ is true in $(D,I)$. 
We use ${\cal I}(D,S)$\ to denote the set of all interpretations for 
the signature $S$\ over the  domain $D$.  

In this paper we are only concerned with finite domains, and 
assume without loss of generality  that $D = D_n:=\{1,\ldots,n\}$\ for some $n\in\Nset$. 

For $I\in {\cal I}(D_n,S)$\ let $\#(i,I)$\ denote the number of elements $\boldd$\ in $D^{k_i}$\ for which 
$I\models\phi_i( \boldd  )$. The \emph{weight} of $I$\ then is
\begin{equation}
\label{eq:semantics1}
  \wnKB(I) := \prod_{i=1}^N w_i^{\#(i,I)},
\end{equation}
where $0^0=1$. The probability of $I$\ is
\begin{displaymath}
  \PnKB(I) = \wnKB(I)/Z
\end{displaymath}
where $Z$\ is the normalizing constant (partition function)
\begin{equation}
  \label{eq:Zconst}
  Z = \sum_{I\in{\cal I}(D_n,S)} \wnKB(I).
\end{equation}

For a first-order sentence $\phi$\ and $n\in\Nset$\  
then 
\begin{equation}
\label{eq:semantics2}
  \PnKB(\phi):= \PnKB( \{ I\in {\cal I}(D_n,S) \mid I\models \phi\} )
\end{equation}
is the probability of $\phi$\ in ${\cal I}(D_n,S)$. 

We call a knowledge base (\ref{eq:kb}) together with the semantics given by 
(\ref{eq:semantics1}) and (\ref{eq:semantics2}) a \emph{weighted feature model},
since it associates weights $w_i$\ with model features $\phi_i$.
Weighted feature models in our sense can be seen as a slight generalization as 
\emph{weighted model counting (wmc)} frameworks~\cite{FieBroThoGutDeR11,GogDom11} 
in which non-zero weights are only associated with literals. Knowledge bases of the form
(\ref{eq:kb}) can be translated into wmc frameworks via an introduction of new 
relation symbols $R_1,\ldots,R_n$, hard constraints $\phi_i(\boldv_i)\leftrightarrow R_i(\boldv_i)$,
and weighted formulas $\wformtxt{R_i(\boldv_i)}{w_i}$~\cite{BroTagMeeDavDeR11,GogDom11}. 
Up to an expansion of the signature, thus, weighted feature models and wmc are 
equally expressive. \emph{Markov Logic Networks}~\cite{RicDom06} also are based on
knowledge bases of the form (\ref{eq:kb}) allowing arbitrary formulas $\phi_i$.
However, the semantics of the model there depends on a transformation of the formulas 
into conjunctive normal form, and therefore does not exactly correspond to 
 (\ref{eq:semantics1}) and (\ref{eq:semantics2}), unless the $\phi_i$\ are clauses.

All types of models here discussed, thus, are very similar in nature, and only differ 
with respect to certain restrictions on what types of logically defined features 
can be associated with a weight. The general definition of weighted feature models gives 
us the flexibility of considering a variety of classes of such restrictions.

A \emph{probabilistic inference problem} $\PInf(\KB,n,\chi,\eta)$\ 
for a weighted feature model is given by a knowledge
base \KB, a domainsize $n\in \Nset$, and two first-order sentences 
$\chi,\eta$. The solution to the inference problem is the conditional probability 
$\PnKB(\chi\mid \eta)$.  

A \emph{class} of inference problems is defined by allowing 
arguments $\KB$, $\chi$, and $\eta$\ only from some restricted classes 
${\cal KB}$, ${\cal Q}$\ (the query class), and ${\cal E}$\ (the evidence class), 
respectively. We use the notation 
\begin{displaymath}
  \PInf({\cal KB},{\cal Q},{\cal E}):=
  \{ \PInf(\KB,n,\phi,\psi)\mid \KB\in {\cal KB}, n\in \Nset, \chi\in{\cal Q},\eta\in {\cal E}     \}
\end{displaymath}
for classes of inference problems.

The results of  this paper will be given for the case where 
${\cal Q}$\ consists of all ground atoms, denoted ${\cal AT}$, and ${\cal E}$\ 
is empty. 
Thus, as far as ${\cal Q}$\ and ${\cal E}$\ are concerned, we are considering 
the most restrictive class of inference problems. Since we are deriving lower
complexity bounds, this leads to the strongest possible results, which directly
apply also to more general classes ${\cal Q}$\ and ${\cal E}$.

Classes ${\cal KB}$\ are defined by various syntactic restrictions on the 
formulas $\phi_i$\ in the knowledge base. In this paper, we consider the following 
fragments of first-order
logic (FOL): relational FOL (RFOL), i.e. FOL without function and constant symbols; 
0-RFOL, which is the quantifier-free fragment of RFOL, and 
0-RFOL$^{\neq}$, which is 0-RFOL without the equality relation. 

An algorithm solves a class $\PInf({\cal KB},{\cal Q},{\cal E})$, if it solves
all instances $\PInf(\KB,n,\chi,\eta)$\ in the class. An algorithm  
$\epsilon$-approximately solves $\PInf({\cal KB},{\cal Q},{\cal E})$, if for 
any \linebreak
$\PInf(\KB,n,\chi,\eta)$\ in the class it returns a number 
$p\in [\PnKB(\chi\mid \eta)-\epsilon,\PnKB(\chi\mid \eta)+\epsilon]$. 
An algorithm that solves $\PInf({\cal KB},{\cal Q},{\cal E})$\ is \emph{polynomial in 
the domainsize}, if for fixed $\KB,\chi,\eta$\ the computation of 
$\PInf(\KB,n,\chi,\eta)$\ is polynomial in $n$.


\section{Spectra and Complexity}
\label{sec:spectra}

The following definition introduces the central concept for our analysis. 

\begin{defenv}
  Let $\phi$\ be a sentence in first-order logic. The \emph{spectrum} of $\phi$\ is
the set of integers $n\in\Nset$\ for which $\phi$\ is satisfiable by an 
interpretation of size $n$.
\end{defenv}

\begin{examenv}
\label{ex:psi1}
  Let $\phi=\psi_1\wedge\psi_2\wedge\psi_3$, where
  \begin{displaymath}
    \begin{array}{lcl}
      \psi_1 & \equiv & \forall x,y\ \ u(x,y) \Leftrightarrow u(y,x) \\
      \psi_2 & \equiv & \forall x\  \exists y\ \  y\neq x \wedge u(x,y) \\
     \psi_3 & \equiv & \forall x,y,y'\ \  (u(x,y) \wedge u(x,y') \Rightarrow y=y' ) 
    \end{array}
  \end{displaymath}
$\phi$\ expresses that the binary relation $u$\ defines an undirected graph 
($\psi_1$) in which every node is connected to exactly one 
other node ($\psi_2,\psi_3$). Thus, $\phi$\ describes a \emph{pairing} relation that 
is satisfiable exactly over domains of even size:   $\spec (\phi)=\{n \mid n \ \mbox{even}\}$. 
\end{examenv}

The complexity class \etime\ consists of problems solvable in time 
$O(2^{cn})$, for some constant $c$. The corresponding nondeterministic class 
is \netime. Note that these classes are distinct from the more commonly 
studied classes (N)EXPTIME, which are characterized by complexity bounds 
$O(2^{n^c})$~\cite{Johnson90}. 
For $n\in\Nset$\ let $\emph{bin}(n)\in \{0,1\}^*$\ denote the binary 
coding of $n$, and $\emph{un}(n)\in \{1\}^*$\ the unary coding (i.e., $n$\ is 
represented as a sequence of $n$\ 1s). 
A set $S\subseteq \Nset$\ is in \nnetime, iff  
$\{\emph{bin}(n)\mid n\in S\}$\ is in \nnetime, which also is equivalent to   
$\{ \emph{un}(n)\mid n\in S\} $\ being in (N)PTIME. 

Like \cite{Jaeger00}, we use the following connection between spectra and 
\netime\ as the key tool for our complexity analysis. 

\begin{theoenv}
 \cite{JonSel72}
\label{theo:jonsel}
A set $ A  \subseteq \Nset$\ is in \netime, iff $A$\ is the spectrum of a 
sentence $\phi\in$\ RFOL. 
\end{theoenv}

\begin{coroenv}
\label{cor:JonSelCor}
  If $\netime \neq \etime$, then there exists a first-order sentence $\phi$,
such that $\{ \emph{un}(n)\mid n\in\spec(\phi)\}$\ is not recognized in 
deterministic polynomial time.
\end{coroenv}

Thus, by reducing  instances $n\in\spec(\phi)?$\ of the spectrum recognition problem to 
probabilistic inference problems $\PInf(\KB,n,\chi,\eta)$, where 
$\KB\in{\cal KB},\chi\in{\cal Q},\eta\in{\cal E}$\ are fixed for the given $\phi$, 
one establishes that the $\PInf({\cal KB},{\cal Q},{\cal E})$\ 
is not polynomial in the domainsize (under the assumption $\etime\neq\netime$).

\section{Complexity Results}
\label{sec:results}

This section contains our complexity results. We begin with a result for knowledge
bases using full RFOL. This is rather straightforward, and (for exact 
inference) already implied by the results of ~\cite{Jaeger00}. We then proceed 
to extend this base result to
0-RFOL and 0-RFOL$^{\neq}$. 

\subsection{Base Result: the RFOL Case}

\begin{theoenv}
\label{theo:RFOL}
  If $\netime \neq \etime$, then there does not exist an algorithm that 0.25-approxi\-mately
solves $\PInf( \mbox{RFOL},  {\cal AT}, \emptyset)$\ in time polynomial in the
domainsize.  
\end{theoenv}


The proof of this theorem provides the general pattern also for subsequent proofs. It is therefore 
here given in full. 

\begin{proofenv}
Let $\phi$\ be a sentence with a non-polynomial spectrum as given by Corollary~\ref{cor:JonSelCor}.
Let $S$\ be the relational signature of $\phi$. Let $a()$\ be a new relation symbol of arity zero
(i.e., $a()$\ represents a propositional variable). The first weighted formula in our knowledge base 
then is 
\begin{equation}
  \label{eq:kb1}
  \wformeq{\neg(\phi \leftrightarrow a())}{0}
\end{equation}
We now already have that $\PnKB(a())>0$\ iff there exists $I\in {\cal I}(D_n,S)$\ with
$I\models\phi$, i.e., iff $n\in\spec(\phi)$. This 
already reduces the decision problem for $\spec(\phi)$\ to solving 
$\PInf(\KB,n,a(),\emptyset)$\ exactly.
However, from the 0-1 laws of first-order logic~\cite{Fagin76}, it 
follows  that for our current \KB: $\PnKB(a())\rightarrow_{n\rightarrow\infty} 0$. 
Thus, for every $\epsilon>0$\ we 
could define an $\epsilon$-approximate constant-time inference algorithm by returning
0 for all sufficiently large $n$. 

In order to obtain our result for approximate inference, we will now ensure that for 
all $n\in\spec(\phi)$\ the probability $\PnKB(a())$\ is greater than 0.5, while it remains
zero for  $n\not\in\spec(\phi)$. We do this 
essentially by calibrating the normalization constant $Z$\ in (\ref{eq:Zconst}). For 
this we introduce another new relation $b()$, and add to \KB: 

\begin{equation}
  \label{eq:kb2}
  \wformeq{\neg ( (\bigwedge _{R\in S} \forall\boldx \neg R(\boldx)) \leftrightarrow b() )}{0}
\end{equation}

Thus, for every $n$\ there is exactly one  interpretation $I\in {\cal I}(D_n,S)$\ 
with nonzero weight in which $b()$\ is true (the one in which all relations have empty interpretations).
Finally, we give zero weight to all interpretations except those in which $a()$\ or $b()$\ is true:
\begin{equation}
  \label{eq:kb3}
    \wformeq{\neg(a()\vee b())}{0}
\end{equation}

Let \KB\ consist of (\ref{eq:kb1}),(\ref{eq:kb2}),(\ref{eq:kb3}). 
Every $I\in {\cal I}(D_n,S)$\ then has weight 0 if it satisfies one of the
three formulas, and weight 1 otherwise.
Consider the case $n\not\in \spec(\phi)$. Then, by (\ref{eq:kb1}) $\wnKB(a())=0$. 
By (\ref{eq:kb3}) this then  means that in all  interpretations of nonzero weight
$b()$\ must be true. By (\ref{eq:kb2}) there is exactly one such interpretation. Thus, 
$Z$\ in (\ref{eq:Zconst}) is 1, and $\PnKB(a())=0/1=0$.

If $n\in \spec(\phi)$, then $\wnKB(a())\geq 1$, and $Z=\wnKB(a())$\ (if the interpretation 
in which all $R$\ are empty also is a model of $\phi$), or $Z=\wnKB(a())+1$\ (otherwise). 
Thus, $\PnKB(a())\geq 1/2$. 
A 0.25-approximate inference algorithm for $\PInf(\KB,n,a(),\emptyset)$, thus, would
decide $\spec(\phi)$. 
\end{proofenv}




\subsection{The 0-RFOL Case}
We now proceed towards our main result, which is going from RFOL to 0-RFOL. 
If we wanted to allow function and constant symbols in our knowledge base, then
one could go to a quantifier-free fragment 
in a quite straightforward manner using Skolemization. Since
satisfiability over a given domain is the same for a formula $\phi$\ and its 
quantifier-free Skolemized version $\phi^{\emph{Skol}}$, the arguments of the
proof of Theorem~\ref{theo:RFOL} would go through with little change.
In order to accomplish the same using only the relational fragment 0-RFOL, 
we define the \emph{relational Skolemization} of a formula. The idea is to 
replace function and constant symbols in the Skolemized version of a formula
with relational representations. For example, the Skolemized version of 
$\psi_2$\ from Example~\ref{ex:psi1} is
\begin{displaymath}
  \psi_2^{\emph{Skol}} \equiv  \forall x \ \ f(x)\neq x \wedge  u(x,f(x)) 
\end{displaymath}
with a new function symbol $f()$. Introducing a relational encoding of $f()$\ leads to
\begin{displaymath}
  \psi_2^{\emph{R-Skol}} \equiv \forall x,y\ R^f(x,y) \rightarrow (y\neq x \wedge u(x,y)) 
\end{displaymath}
with $R^f$\ a new binary relation symbol encoding $f()$. 
This translation must be accompanied by axioms that 
confine the possible interpretations
of $R^f$\ to relations that encode functions.

Such relational encodings of functions are well established. However, there does not seem
to be a standard account of this technique that serves our purpose. The 
following proposition, therefore, provides the relevant result in a form tailored 
for our needs.

\begin{propenv}
\label{prop:r-skol}
  Let $\phi(\boldx)\in\mbox{0-FOL}(S\cup S^F)$, where $S$\ is  a set of relation symbols, and 
$S^F$\ a set of function and constant symbols. Let $S^+$\ be a set of new relation
symbols that for every $k$-ary $f\in S^F$\ contains a $k+1$-ary $R^f$\ (constant symbols
are treated as 0-ary function symbols). Let \emph{Func} be the set of sentences that 
for every $f\in S^F$\ contains
\begin{eqnarray}
  & \forall \boldx\, y\, y' \ (R^f(\boldx,y)\wedge R^f(\boldx,y') \rightarrow y=y'   ) 
  &  \label{eq:FuncAx1} \\
    & \forall \boldx \exists y \  R^f(\boldx,y).
  &  \label{eq:FuncAx2}
\end{eqnarray}
Then there exists a formula $\phi^+(\boldx,\boldz)\in $0-RFOL$(S\cup S^+)$, such that 
the following are equivalent for all $n$:
\begin{description}
\item[i] there exists $I\in{\cal I}(D_n,S\cup S^F)$\ with $I\models \forall\boldx \phi(\boldx)$
\item[ii] there exists $I^+\in {\cal I}(D_n,S\cup S^+)$\ with
$I^+\models \emph{Func} \wedge \forall \boldx\boldz\, \phi^+(\boldx,\boldz)$
\end{description}
\end{propenv}

If $\phi^{\emph{Skol}}$\ 
is the Skolemization of a formula $\phi\in$RFOL, we then call ${ \phi^{\emph{Skol}}}^+$\ the
\emph{relational Skolemization} of $\phi$, written $\phirskol$.  

Our plan, now, is to prove the analogon of Theorem~\ref{theo:RFOL} for 
0-RFOL by replacing $\phi$\ in (\ref{eq:kb1}) with $\phirskol$. However, this
is not enough, since we also 
need to constrain the models of our knowledge base (more precisely: those models in which 
$a()$\ is true) to satisfy the axioms (\ref{eq:FuncAx1}) and
(\ref{eq:FuncAx2}). This poses a problem, because (\ref{eq:FuncAx2}) contains an 
existential quantifier, and so we cannot add this axiom directly as a constraint 
to a knowledge base restricted to 0-RFOL. Indeed, we almost seem to have gone full circle, 
since we are back at knowledge bases in a relational vocabulary with existential quantification!
However,  we now have 
reduced arbitrary occurrences of existential quantifiers to occurrences only
within in the special formulas (\ref{eq:FuncAx2}).

Our strategy, now, is  to approximate formulas (\ref{eq:FuncAx2}) with
weighted formulas of the form
\begin{equation}
\label{eq:approxexist}
  \wformeq{a() \wedge R^f(\boldx,y)}{w}
\end{equation}
that reward models of $a()$\ in which the existential quantifier of (\ref{eq:FuncAx2}) 
is satisfied for many (all) $\boldx$. We will no longer be able to ensure that
$\wnKB(a())=0$\ when $n\not\in\spec (\phi)$. However, by a suitable choice of $w$,
and by a careful calibration of the weight of models of the alternative proposition
$b()$, we still can ensure that $\wnKB(b())\gg \wnKB(a())$\ when 
$n\not\in\spec (\phi)$, and $\wnKB(b())\approx \wnKB(a())$\ when 
$n\in\spec (\phi)$.  However, the right calibration of the weights of models of $a()$\ and $b()$\
within ${\cal I}(D_n,S)$\ will now require that one sets $w$\ to a value $w(n)$\ depending
on $n$.

This means that we no longer can reduce the decision problem  $n\in\spec(\phi)$\ to the 
probabilistic inference problem $\PInf(\KB,n,a(),\emptyset)$\ for a fixed 
knowledge base $\KB$. We only achieve a reduction  to the  inference problem 
$\PInf(\KB(w(n)),n,a(),\emptyset)$, where the logical structure of $\KB$\ is fixed, 
but a weight parameter $w(n)$\ depends on $n$. 
Generally, for a knowledge base $\KB$\ containing $N$\ weighted formulas, 
we denote with $\KB(w_1,\ldots,w_N)$\ the knowledge base that contains the
same formulas as $\KB$, but with the weights set to values $w_1,\ldots,w_N$. 

To translate the lower complexity 
bounds of the original spectrum recognition problem into lower complexity bounds for the resulting
inference problem, one now has to be precise about the representation of the 
inference problem. To this end, we assume that weights $w$\ are rational numbers, and 
represented by pairs $(u,v)$\ of integers, so that  $w=u/v$.  We then define the representation size
$l(w)$\ as $\log(\cardinal{u}+1)+\log(\cardinal{v}+1)$.
The total representation size of the weight parameters $\boldw=(w_1,\ldots,w_N)$\ in a knowledge base
is $l(\boldw):= \sum_{i=1}^N l(w_i)$. An inference algorithm for probabilistic inference 
problems in $\PInf({\cal KB},{\cal Q},{\cal E})$\ is \emph{polynomial in the domainsize and 
the representation size of the weight parameters}, if for any 
$\KB\in {\cal KB}$, $\chi\in {\cal Q}$, $\eta\in {\cal E}$\ the class of inference problems
$\PInf(\KB(\boldw),n,\chi,\eta)$\ can be solved in time that is bounded by a polynomial
$\sum_{i,j=0}^d \alpha_{i,j} l(\boldw)^i n^j$\ ($\alpha_{i,j}\in\Rset, d\in\Nset  $ ). 
We can now state the following theorem:

\begin{theoenv}
\label{theo:main1}
  If $\netime \neq \etime$, then there does not exist an algorithm that 0.2-approximate\-ly solves
$\PInf( \mbox{0-RFOL}, {\cal AT}, \emptyset)$\ in time polynomial in the domainsize
and  the representation size of the weight parameters.  
\end{theoenv}

The full proof of the theorem is given in the appendix.
It consists of a polynomial-time reduction of the $n\in\spec(\phi)$\ decision problem to a probabilistic
inference problem $\PInf(\KB(\boldw(n)),n, a(),\emptyset)$, where $l(\boldw(n))$\ is polynomial in 
$n$. An inference algorithm that can solve  $\PInf(\KB(\boldw(n)),n, a(),\emptyset)$\ in time 
polynomial in the domainsize and $l(\boldw(n))$, thus, would yield a polynomial decision procedure
for $\spec(\phi)$.  

\subsection{Polynomiality in $l(\boldw)$}

One may  wonder how strong or surprising Theorem~\ref{theo:main1}
really is in light of its extra  runtime polynomial in $l(\boldw)$\ condition. 
It has previously been emphasized that lifted inference procedures should only be expected to be 
polynomial in the domain size, but not in other parameters that characterize the complexity
of \KB~\cite{Jaeger00,Broeck11}.  These remarks, however, have mostly been 
motivated by considerations
of the logical complexity of \KB, e.g. in terms of the number and complexity of its weighted 
formulas, or the size of the signature. 
The complexity in terms of numerical parameters, on the other hand, has not received
much attention.

To better understand the nature of the condition of being  polynomial in
the domainsize and  $l(\boldw)$, we have to look a little closer at how the 
parameters affect the complexity of the computation. 
We consider algorithms that can be described as follows: 
to compute $\PInf(\KB(\boldw),n,\chi,\eta)$\ the algorithm performs
a number of steps $i=1,\ldots, L$,   where step $i$\ consists  either of 
executing a constant time operation that does not depend on the numerical model parameters
(e.g., a logical operation on formulas), or of a basic operation
on numerical parameters. 

We consider the executions the algorithm performs on inputs with fixed logical 
structure $\KB$, and fixed $\chi,\eta$, but varying weight parameters $\boldw$\ and
domainsizes $n$. 
Let $V_{\boldw,n}(i)$\ denote the  set of all numerical variables stored by the algorithm 
before performing step $i$, when it is run on inputs $(\boldw,n)$. 
Thus, $V_{\boldw,n}(i)$\ comprises the original weight parameters of the model,
as well as computed intermediate results, etc. 
We now make two basic assumptions on the algorithm:
\begin{description}
\item[(A1)] The weight parameters $\boldw$\ only influence the numerical values
of the variables stored in 
$V_{\boldw,n}(i)$, but not the sequence of execution steps performed 
by the algorithm.  In particular, the number of execution steps performed by the 
algorithm only depends on $n$: $L=L(n)$. 
\item[(A2)] The basic operations
performed on numerical variables are polynomial time in the size of their 
arguments, and  they produce an output whose size is linear in the size of the inputs.
This is the case for the basic arithmetic operations addition and multiplication, 
for example.
\end{description}

The total representation size of $V_{\boldw,n}(i)$\ then is bounded by 
$c_n(i) l(\boldw)$, where $c_n(i)$\ is a coefficient not depending on $\boldw$.
Also, let $q()$\ be a polynomial that provides a common complexity bound 
for the basic numerical operations that can be performed at one step.   
The total
execution time of the algorithm on input $(\boldw,n)$\ then is bounded by 
\begin{equation}
\label{eq:runtime}
  \sum_{i=1}^{L(n)}    q (c_n(i) l(\boldw)).
\end{equation}

If, now, for fixed weight vectors $\boldw$\ the algorithm is polynomial in $n$\ 
(equivalently: the algorithm is polynomial in $n$\ under a 
computation model where basic numeric 
operations are constant time), then $L(n)$\ and  
$\max_{i=1,\ldots,L(n)} c_n(i)$\ must be  polynomially
bounded in $n$. The combined complexity (\ref{eq:runtime}) then, in fact,  is 
polynomial both in $n$\ and $l(\boldw)$.

In summary, this shows: an algorithm that for fixed $\boldw$\ is polynomial in 
$n$, and that satisfies assumptions (A1) and (A2), actually is polynomial
in $n$\ and $l(\boldw)$. Thus, for this type of algorithm, the additional 
restriction of Theorem~\ref{theo:main1} compared to Theorem~\ref{theo:RFOL}
is insignificant. 

The remaining question, then, is how restrictive or realistic assumptions 
(A1) and (A2) actually are. For exact inference algorithms it appears that (A1) and (A2) 
are satisfied by all existing approaches, with a small qualification: algorithms
might give special treatment to special weight parameters, such as $w=0$\ or
$w=\infty$, which then can lead to a  violation of (A1) in the strict sense. However, our analysis 
could also be performed based on a weakened form of (A1) that allows certain special weights
to influence the computation differently from proper numerical weights
$0<w<\infty$. A slightly more elaborate argument would then arrive at essentially
the same conclusions. 

The situation is less clear for approximate inference algorithms. Here the numerical 
values stored in $V_{\boldw,n}(i)$\ may influence the algorithm in multiple ways: 
for example, they can be used to test a termination condition, or to decided which computations to 
perform next in order to improve approximation bounds derived so far. In all such cases, 
the model weights $\boldw$\  can have an impact on the sequence and the total number of execution
steps, and (A1) is not satisfied. Thus, even though the theorem  also applies to 
approximate inference, its implications for the construction of approximate inference algorithms
may be less severe, since there might be reasonable ways to build approximate inference algorithms that 
are polynomial in $n$, without also being polynomial in $l(\boldw)$.  

\subsection{The $\mbox{0-RFOL}^{\neq}$\ Case}
In a final strengthening of our results, we now move on to the fragment
$\mbox{0-RFOL}^{\neq}$. The availability of the equality predicate for the 
formulas of \KB, so far, has been an important prerequisite for our arguments, because
Theorem~\ref{theo:jonsel} crucially depends on equality: spectra for formulas
$\phi\in \mbox{RFOL}^{\neq}$\ are always of the form $\Nset\setminus\{1,\ldots,k\}$\ for 
some $k$, and, thus, decidable in constant time. For this reason it was suggested 
in \cite{Jaeger00}  that one should focus on logical fragments without equality
when looking for model classes for which  lifted inference scales polynomially in 
the domainsize. As our 
final result shows, however, elimination of equality may not have such a large 
impact on complexity, after all.

\begin{theoenv}
\label{theo:main2}
  If $\netime \neq \etime$, then there does not exist an algorithm that 0.2-approximate\-ly solves
$\PInf( \mbox{0-RFOL}^{\neq}, {\cal AT}, \emptyset)$\ in time polynomial both 
in the domainsize, and  the representation size of the weight parameters. 
\end{theoenv}

This theorem is a generalization of Theorem~\ref{theo:main1}, and, strictly speaking, 
makes \ref{theo:main1} redundant. It is only for expository purposes, and greater 
transparency in the proof arguments, that we here develop these results in two 
steps.

The proof of Theorem~\ref{theo:main2} is a refinement of the proof of Theorem~\ref{theo:main1}.
In addition to approximating Skolem functions $f$\ with relations $R^f$,  
we now also approximate the equality predicate 
$ = $\ with a binary relation $E(\cdot,\cdot)$. Similarly as we could not impose in 
0-RFOL hard constraints that ensure that $R^f$\ encodes a function, we also cannot 
constrain models to always interpret $E$\ as the equality relation. However, just as 
with  (\ref{eq:FuncAx1})  and (\ref{eq:approxexist})  we rewarded interpretations
with functional  $R^f$, we can penalize interpretations in which $E$\ is not  true equality 
by means of the two weighted formulas
\begin{eqnarray}
  & & \wformeq{a()\wedge \neg E(x,x)}{0} \label{eq:approxE1}\\
   & & \wformeq{a()\wedge E(x,y)}{1/w} \label{eq:approxE2}
\end{eqnarray}
where $w$\ is a large weight. 

\section{Approximate Inference ,  Convergence, and Evidence}
\label{sec:discuss}

There are some notable differences with respect to approximate inference 
between the results we here obtained for 
weighted model counting, and the results of \cite{Jaeger00}.
In  \cite{Jaeger00} it was shown that 
due to convergence of query probabilities $P_n(a())$\ 
as $n\rightarrow\infty$, in theory a trivial 
constant time approximation algorithm exists: perform exact inference 
for all input domains up to a size $n^*$, and output the limit probability for all 
domains of size $ > n^*$. This ``algorithm'', however, has no practical use, since for a desired 
accuracy value $\epsilon$\ one first would have to determine a sufficiently high threshold value 
$n^*\in\Nset$\ to make the output indeed be an $\epsilon$-approximation. 

Nevertheless, the difference between the existence of an impractical approximation algorithm on 
the one hand, and the non-existence of any approximation algorithm on the other hand, is just one
consequence of a more fundamental difference: while in the models considered in \cite{Jaeger00}
query probabilities $P_n(a())$\ converge to a limit, this is not necessarily the case  for knowledge bases 
of weighted formulas -- at least when full RFOL is allowed: in the proof of
Theorem~\ref{theo:RFOL}  we have  constructed knowledge bases \KB, such that 
$\PnKB(a())$\ oscillates between  zero 
and values $>1/2$\ as $n$\ oscillates between $\spec(\phi)$\ and
its complement. The construction of knowledge bases with this behavior does not require 
formulas $\phi$\ with a non-polynomial spectrum as in Corollary~\ref{cor:JonSelCor}, 
and is not contingent on $\netime\neq\etime$. Already a knowledge base as constructed  in 
the proof of Theorem~\ref{theo:RFOL} with $\phi$\ replaced by $\psi$\ of Example~\ref{ex:psi1} 
will show this behavior. 

The reason behind these different convergence properties lies in a somewhat different role that
conditioning on evidence plays in directed and undirected models: in the former, 
a conditional probability $\PnM(a()\mid b())$\ defined by a model $M$\ can, in general, 
not be defined as an unconditional probability $\PnMpr(a())$\ in a modified model $M'$. 
As a result, the convergence guarantees and -- theoretical -- approximability 
for certain classes of unconditional queries $\PnM(a())$, do not carry over to 
conditional queries $\PnM(a()\mid b())$. 

For weighted feature knowledge bases $\KB$, on the other hand, there is no fundamental 
difference between unconditional and conditional queries 
$\PnKBpr(a())$\ and $\PnKB(a()\mid b())$,    respectively. 
To reduce the conditional to unconditional queries, one can just add to 
$\KB$\ the hard constraint 
$\wformtxt{\neg b()}{0}$\ to obtain $\KB'$\ with $P\PnKBpr=\PnKB \mid b()$. 
This means that as long as ${\cal E}$\ is not more expressive than
${\cal KB}$, the problem classes $\PInf({\cal KB},{\cal Q},{\cal E})$\ and 
$\PInf({\cal KB},{\cal Q},\emptyset)$\ have the same characteristics in terms of complexity
as a function of the domainsize. Note, though, that this is only true when
we consider complexity of $\PInf(\KB,n,\chi,\eta)$\ strictly as a function of $n$\ for fixed 
$\KB,\chi,\eta$.  If the evidence is allowed to change with the domainsize, i.e., 
$\eta=\eta(n)$, then even in cases where restrictions on ${\cal KB}$\ make 
$\PInf({\cal KB},{\cal Q},{\cal E})$\ polynomial in $n$, one can define 
sequences of inference problems $\PInf(\KB,n,\chi,\eta(n))$\ with 
$\KB\in{\cal KB}$, $\eta(n)\in {\cal E}$\ that are no longer 
polynomial in $n$~\cite{Broeck12}.

\section{Conclusion}

We have shown that for currently quite popular  probabilistic-logic models consisting 
of collections of weighted, quantifier- and function-free formulas there is likely to be no 
general polynomial lifted inference method (contingent on  $\netime\neq\etime$). 
Somewhat surprisingly, this even holds for approximate inference. Between this negative result, and 
the positive result of \cite{Broeck11}, there still could be a lot of room for identifying tractable 
fragments by restricting 0-RFOL further via limits on the number of variables, or the richness of the 
signature $S$.

\appendix

\section{Proofs}

\begin{proofofenv}{of Proposition~\ref{prop:r-skol}}

  We begin by defining the \emph{term-depth} of a term $t$\ in the signature $S^F$\ as the maximal 
nesting depth of function symbols in $t$. Precisely, we define inductively: if $t\equiv x$, then
$t$\ has term depth 0. If $t\equiv f()$\ (a constant), or $t=f(x_1,\ldots,x_k)$\ (a function term
with only variables as arguments), then $t$\ has term depth 1. If $t=f(t_1,\ldots,t_k)$, then 
the term depth of $t$\ is one plus the maximal term depth of the $t_i$. 

The term depth of a formula $\phi(\boldx)$\ is the maximal term depth of the terms it contains. 

We now show that every formula $\phi(\boldx)$\ of term depth $l$\ can be transformed into 
a formula $\phi^{l-1}(\boldx,\boldz)$\ of term depth $l-1$\ in 0-FOL$(S\cup S^F \cup S^+)$,  
such that the statement for $\phi^+$\ of the proposition holds for $\phi^{l-1}$\ (but with 
$S\cup S^F\cup S^+$\ instead of $S\cup S^+$ in {\bf ii}). The proposition
then follows by defining $\phi^+$\ as the result of iteratively applying  $l$\ such transformations
to $\phi$. Since the term depth of the resulting $\phi^+$\  is zero, then 
actually $\phi^+(\boldx,\boldz)\in $0-RFOL$(S\cup S^+)$.

Let $\{f_i(\boldx_i)\mid i=1,\ldots,r\}$\ be the set of all distinct terms (including sub-terms) 
of depth 1 appearing in $\phi(\boldx)$. Let $z_1,\ldots,z_r$\ be new variables. Define 
$\phi^{l-1}(\boldx,\boldz)$\ as
\begin{displaymath}
  \bigwedge_{i=1}^r R^{f_i}(\boldx_i,z_i) \rightarrow 
  \phi(\boldx)[z_1 / f_1(\boldx_1) , \ldots, z_r / f_r(\boldx_r)   ]
\end{displaymath}

To now show {\bf i}$\Rightarrow${\bf ii} let $I\in{\cal I}(n,S\cup S^F)$\ with 
$I\models \forall\boldx \phi(\boldx)$. Define $I^+\in {\cal I}(n,S\cup S^F\cup S^+)$\ 
as the expansion of $I$\ in which 
each $R^f\in S^+$\ is interpreted as the relational representation of $f$, i.e., 
$I^+ \models R^f(\boldd,e)$\ iff $I\models f(\boldd)=e$. Clearly, $I^+\models \emph{Func}$. 
Furthermore, the following are equivalent:
\begin{displaymath}
\begin{array}{l}
  I\models \forall\boldx\, \phi(\boldx) \\
  I\models \forall\boldx\boldz \bigwedge_{i=1}^r f_i(\boldx_i)=z_i \\
  \hspace{10mm} \rightarrow 
  \phi(\boldx)[z_1 / f_1(\boldx_1)  , \ldots, z_r /  f_r(\boldx_r)   ] \\
  I^+\models \forall\boldx\boldz \bigwedge_{i=1}^r R^{f_i}(\boldx_i,z_i)\\
  \hspace{10mm} \rightarrow 
  \phi(\boldx)[z_1 / f_1(\boldx_1)  , \ldots, z_r / f_r(\boldx_r)   ] \\
\end{array}
\end{displaymath}

For {\bf ii}$\Rightarrow${\bf i} let $I^+$\ as in {\bf ii} be given. Since $I^+\models\emph{Func}$,
we can turn $I^+$\ into an interpretation for $S\cup S^F$\ by defining 
$f(\boldd)$\ as the unique $e$\ for which $R^f(\boldd,e)$\ holds in $I^+$. Then, by the 
same equivalences as above, $I^+\models \forall \boldx\boldz\, \phi^+(\boldx,\boldz)$\ 
implies $I\models \forall\boldx \phi(\boldx)$.
\end{proofofenv}

\begin{proofofenv}{of Theorem~\ref{theo:main1}}
Let $\phi\in$RFOL as given by Corollary~\ref{cor:JonSelCor}, and $\forall\boldx\ \phirskol(\boldx)$\ 
its relational 
Skolemization. Let $S$\ be the original signature of $\phi$, and $S^+$\ 
the relation symbols introduced in the relational Skolemization.
Furthermore, for each $k$-ary $R^+\in S^+$\ we introduce a new 
$(k-1)$-ary relation $R^{++}$. These new symbols will be used to calibrate the
weight of models for the reference proposition $b()$. 
Note that the arity of symbols in 
$S^+$\ is at least 1, and $R^{++}$, thus, is well-defined, but may  contain 
relations of arity 0. We denote with $S^{++}$\ the collection of all the introduced $R^{++}$\ symbols.  
We now reduce the spectrum recognition problem for $\phi$\ 
to probabilistic inference from a knowledge base in the signature
$S\cup S^+ \cup S^{++} \cup \{ a(), b()\}$. 

The first formula in our knowledge base is
\begin{equation}
\label{eq:kb21}
 \wformeq{a() \wedge \neg \phirskol(\boldx)}{0}
\end{equation}

We now approximately axiomatize the functional nature of the 
symbols $R^+\in S^+$. The sentence (\ref{eq:FuncAx1}) can be directly
encoded as a weighted formula:

\begin{equation}
  \label{eq:kb22}
  \wformeq{R^+(\boldx,y)\wedge R^+(\boldx,y') \wedge y\neq y'}{0}
\end{equation}

Next, we would like to enforce (\ref{eq:FuncAx2}) by means of a weighted 
formula. However, (\ref{eq:FuncAx2}) encodes  the essence of the existential 
quantifiers we are about to eliminate, and, thus, it is not surprising that this
is not possible to enforce strictly. However, we can reward models 
in which the existential quantification of (\ref{eq:FuncAx2}) is satisfied  
via the weighted formulas

\begin{equation}
  \label{eq:kb23}
  \wformeq{a() \wedge R^+(\boldx,y)}{w} \hspace{5mm} (R^+\in S^+)
\end{equation}
where $w>1$\ is a weight whose exact value is to be defined later. 

We now proceed with constraining models of the reference proposition $b()$. First,  
all symbols in $S\cup S^+$\ shall have an empty interpretations in models of $b()$:
\begin{equation}
  \label{eq:kb24}
  \wformeq{b() \wedge  R(\boldx)}{0}  \hspace{5mm} (R\in S)
\end{equation}
\begin{equation}
  \label{eq:kb25}
  \wformeq{b() \wedge  R^+(\boldx,y)}{0} \hspace{5mm} (R^+\in S^+)
\end{equation}

In order to allow $b()$-models to gain some weight, we use the 
extra symbols in $S^{++}$:

\begin{equation}
  \label{eq:kb26}
  \wformeq{b() \wedge  R^{++}(\boldx)}{w} \hspace{5mm} (R^{++}\in S^{++})
\end{equation}
where $w$\ is the same weight as in (\ref{eq:kb23}). 
To further limit the possible interpretations of $b()$-models, we also stipulate:
\begin{equation}
  \label{eq:kb26b}
  \wformeq{b() \wedge  \neg R^{++}(\boldx)}{0} \hspace{5mm} (R^{++}\in S^{++})
\end{equation}

The extra symbols $R^{++}$\  must have empty interpretations in $a()$-models:
\begin{equation}
  \label{eq:kb27}
  \wformeq{a() \wedge  R^{++}(\boldx)}{0} \hspace{5mm} (R^{++}\in S^{++})
\end{equation}

Finally, we add:
\begin{equation}
  \label{eq:kb28}
  \wformeq{\neg( a()\vee b())}{0}
\end{equation}

We now determine (approximately) $\wnKB(a())$\ and $\wnKB(b())$\ for the cases 
$n\in\spec(\phi)$\ and $n\not\in\spec(\phi)$.

First, consider $b()$: for  any $n$, there exists exactly one interpretation 
$I_{b()}\in {\cal I}(D_n, S\cup S^+ \cup S^{++} \cup \{ a(),b()\})$\ 
with nonzero weight in which $b()$\ is true. This is the 
interpretation in which all relations in $S\cup S^+$\ are empty 
((\ref{eq:kb24}),(\ref{eq:kb25})), all relations 
in $S^{++}$\ are maximal (\ref{eq:kb26b}), and, in consequence of the latter, because 
of  (\ref{eq:kb27}), $a()$\ is false. 

Assume that $S^+ =\{ R^+_1,\ldots,R^+_m \}$, where $R^+_i$\ has arity $k_i+1$. Then
$R^{++}_i\in S^{++}$\ contributes  via (\ref{eq:kb26}) 
a factor of $w^{n^{k_i}}$\ to $\wnKB(I_{b()})$, and the total weight is:
\begin{equation}
  \label{eq:weightofb}
  \wnKB(I_{b()}) = \wnKB(b()) = w^{n^{k_1} +\cdots + n^{k_m}} = w^{K(n)}, 
\end{equation}
using for abbreviation  $K(n):= n^{k_1} +\cdots + n^{k_m}$.

We next turn to $\wnKB(a())$\ in the case $n\in\spec(\phi)$. Then there exists at 
least one interpretation $I\in {\cal I}(D_n,S\cup S^+)$, in which $\forall\boldx\,\phirskol(\boldx)$\  
is true, and in which 
the relations from $S^+$\ have a functional interpretation. 
We can expand this interpretation to an interpretation  in 
${\cal I}(n,S\cup S^+ \cup  S^{++} \cup \{ a(),b()\})$\ by giving all 
relations in $S^{++}$\ an empty interpretation, and setting $a()$\ to true and
$b()$\ to false. Then $I$\ does not violate any hard constraint in \KB, and 
collects from (\ref{eq:kb23}) a total weight of $w^{K(n)}$. Thus
\begin{displaymath}
  \wnKB(a()) \geq w^{K(n)}, 
\end{displaymath}
and therefore, when $n\in\spec(\phi)$
\begin{equation}
  \label{eq:pofa}
  \PnKB(a())\geq \wnKB(a())/(\wnKB(a())+\wnKB(b())) \geq 1/2.
\end{equation}

Finally, we have to consider $\wnKB(a())$\ in the case $n\not\in\spec(\phi)$.
For any $I$\ with nonzero weight in which $a()$\ is true, because of (\ref{eq:kb21}),
also $\forall\boldx\phirskol(\boldx)$\ must be true. This, now, only is possible when some $R^+\in S^+$\ is 
not a functional relation, which, because of (\ref{eq:kb22}) can only mean that for some 
$\boldx$\ there exists no $y$\ with $R^+(\boldx,y)$. The total weight of $I$\ accrued from
(\ref{eq:kb23}) then is at most $w^{K(n)-1}$. Because of (\ref{eq:kb27}), $I$\ cannot 
obtain any additional weight from (\ref{eq:kb26}), so that
\begin{equation}
  \label{eq:WIbound}
  \wnKB(I)\leq w^{K(n)-1}.
\end{equation}
The total number of interpretations in ${\cal I}(D_n,S\cup S^+ \cup  S^{++} \cup \{ a(),b()\})$\ is
$2^{L(n)}$\ for a polynomial $L(n)$. Thus
\begin{equation}
  \label{eq:Wabound}
  \wnKB(a())\leq 2^{L(n)} w^{K(n)-1}.
\end{equation}

We now obtain for the case  $n\not\in\spec(\phi)$
\begin{equation}
\label{eq:Pofabound}
  \PnKB(a())  \leq  \wnKB(a())/ \wnKB(b()) 
   \leq    2^{L(n)} w^{K(n)-1}/ w^{K(n)} 
   =  2^{L(n)}/w.
\end{equation}

Setting $w =  10\cdot 2^{L(n)}$, we thus have $\PnKB(a())\leq 1/10$\ if $n\not\in\spec(\phi)$.
The representation size of $w$\ is polynomial in $n$. Thus, an algorithm that 
computes $\PnKB(a())$\ up to an accuracy of $ 0.2 = (0.5-0.1)/2$\ in time polynomial
in $n$\ and the representation size of $w$\ would give a polynomial time decision 
procedure for $\spec(\phi)$. 
\end{proofofenv}

 \begin{proofofenv}{of Theorem~\ref{theo:main2}}
   The proof is an extension of the proof of Theorem~\ref{theo:main1}, and we here just 
 give the necessary modifications. 

 Let $E$\ be a new binary relation symbol. We replace equalities $x=y$\ in 
 (\ref{eq:kb21}) and (\ref{eq:kb22}) with $E(x,y)$. To (approximately) axiomatize
 $E$\ as the identity relation in models of $a()$, we add to the knowledge base
 consisting of (\ref{eq:kb21})-(\ref{eq:kb28}) the weighted formulas 
 \begin{eqnarray}
   & a()\wedge \neg E(x,x) &  0  \label{eq:E1} \\
   & a()\wedge  E(x,y) &  1/w  \label{eq:E2} 
 \end{eqnarray}
 where $w>1$\ is the same weight as in (\ref{eq:kb23}) and (\ref{eq:kb26}), and whose
 exact value is to be determined later. 
 To calibrate the weight of $b()$-models, we introduce in analogy to the 
 $R^{++}$\ relations a unary relation $E^{++}$, and in analogy to 
 (\ref{eq:kb26}) - (\ref{eq:kb27}) add to the knowledge base
 \begin{eqnarray}
   & b()\wedge E^{++}(x) &  1/w  \label{eq:E3} \\
   & b()\wedge \neg E^{++}(x) &  0  \label{eq:E4} \\
   & a()\wedge  E^{++}(x) &  0 \label{eq:E5} 
 \end{eqnarray} 

 We now obtain for all $n$
 \begin{equation}
   \label{eq:local30}
   \wnKB(b())=w^{K(n)}(1/w)^n = w^{K(n)-n}.
 \end{equation}

 If $n\in\spec(\phi)$, then there exists an interpretation in which 
 $a()$\ is true, the $R^+$\ have a functional interpretation, and 
 the interpretation of $E$\ is the identity relation. We can thus lower-bound
 the weight of $a()$\ by the weight of that interpretation:
 \begin{equation}
   \label{eq:local32}
   \wnKB(a())\geq w^{K(n)}(1/w)^n  = w^{K(n)-n}.
 \end{equation}
As in (\ref{eq:pofa}), one then obtains
$\PnKB(a())\geq 1/2$.
  
We now turn to the case $n\not\in\spec(\phi)$. Consider any $I$\ in which
$a()$\ is true, and that has nonzero weight. This now, only is possible 
when in $I$\ there is an $R^+\in S^+$\ which is 
not a functional relation, or when $E$\ is not the identity relation in $I$\
(or both). In all  cases, the weight of $I$\ coming from (\ref{eq:kb23}) and
(\ref{eq:E2}) is at most  $w^{K(n)-n-1}$. 
The total number of interpretations in 
${\cal I}(D_n,S\cup S^+ \cup  S^{++} \cup \{ a(),b(),E\})$\ is
$2^{M(n)}$\ for a polynomial $M(n)$. Thus
\begin{equation}
  \label{eq:Waboundii}
  \wnKB(a())\leq 2^{M(n)} w^{K(n)-n-1},
\end{equation}
from which, as in (\ref{eq:Pofabound}), then $\PnKB(a())  \leq   2^{M(n)}/w$.
Now setting $w=10\cdot 2^{M(n)}$\ again yields the bound
$\PnKB(a())\leq 1/10$.

 \end{proofofenv}

\bibliography{ref,mypubs}

\end{document}